\documentclass[letterpaper, 10 pt, conference]{ieeeconf}  

\IEEEoverridecommandlockouts                              

\usepackage{graphics} 
\usepackage{epsfig} 
\usepackage{mathptmx} 
\usepackage{times} 
\usepackage{microtype}
\usepackage{algorithmic}
\usepackage{textcomp}
\usepackage{xcolor}
\usepackage{gensymb}
\usepackage{caption}
\usepackage{hyperref}
\usepackage{mwe}
\usepackage{verbatim}
\usepackage{amsbsy}
\usepackage{siunitx}
\usepackage{tabularx,booktabs}
\usepackage{dblfloatfix}
\usepackage{cite}
\usepackage{bm}
\usepackage{amsmath,amssymb,amsfonts}
\usepackage{subcaption}
\usepackage{float}

\title{\LARGE \bf
Jubileo: An Open-Source Robot and Framework\\for Research in Human-Robot Social Interaction
}

\author{Jair A. Bottega$^{1}$, Victor A. Kich$^{1}$, Alisson H. Kolling$^{1}$, Jardel D. S. Dyonisio$^{2}$,\\Pedro L. Corçaque$^{2}$, Rodrigo da S. Guerra$^{2}$, Daniel F. T. Gamarra$^{1}$
\thanks{$^{1}$Jair A. Bottega, Victor A. Kich, Alisson H. Kolling and Daniel F. T. Gamarra are with Federal University of Santa Maria - UFSM, Santa Maria, RS, Brazil.
        {\tt\small jairaugustobottega@gmail.com}}%
\thanks{$^{2}$Jardel D. S. Dyonisio, Pedro L. Corçaque and Rodrigo da S. Guerra are with the NAUTEC, Centro de Ciencias Computacionais, Universidade Federal do Rio Grande - FURG, RS, Brazil.
        {\tt\small tioguerra@gmail.com}}%
}

\begin{document}

\maketitle
\thispagestyle{empty}
\pagestyle{empty}

\begin{abstract}


Human-robot interaction (HRI) is essential to the widespread use of robots in daily life. Robots will eventually be able to carry out a variety of duties in human civilization through effective social interaction. Creating straightforward and understandable interfaces to engage with robots as they start to proliferate in the personal workspace is essential. Typically, interactions with simulated robots are displayed on screens. A more appealing alternative is virtual reality (VR), which gives visual cues more like those seen in the real world. In this study, we introduce Jubileo, a robotic animatronic face with various tools for research and application development in human-robot social interaction field. Jubileo project offers more than just a fully functional open-source physical robot; it also gives a comprehensive framework to operate with a VR interface, enabling an immersive environment for HRI application tests and noticeably better deployment speed.

\end{abstract}

\section*{SUPPLEMENTARY MATERIAL}

Video of the real and virtual robot working using its framework is available at \url{https://youtu.be/JuxAU4nFGbk}. Released code, Docker image, and 3D models at \url{https://github.com/jajaguto/jubileo}.

\section{INTRODUCTION}\label{introduction}

The emergence of robots in society has driven the research of Human-Robot Interaction (HRI) to produce intelligent robots that can serve in the human living environment. Goodrich and Schultz \cite{goodrich} define HRI as a field of study dedicated to understanding, designing, and evaluating robotic systems for use with humans. Human-Robot social interaction requires communication between robots and humans, which can be accomplished in several ways. Humans communicate with their voice, gestures, and body language and rely on facial expressions to convey our feelings and attitudes to others and interpret other people's emotions, desires, and intentions \cite{plutchik1984emotions}. In order to achieve direct social interaction between a human and robot, humanoid robots possessing human-like capabilities and replicating the human figure were developed.  

This work presents Jubileo, an open-source animatronic face, which includes a complete operational system and simulation framework for researchers in the HRI field to develop social applications. Furthermore, the robot face is intended to escape from the Uncanny Valley \cite{mori2012uncanny} concept, which is a state where an object's appearance is very similar to a human's, resulting in a negative emotional response to it. Thus, Jubileo's face has a friendly appearance designed to minimize discomfort and encourage humans to engage in a conversation, where the robot can react socially, processing and returning an answer by voice, gesture, or facial expressions.

Since its inception, simulation in the field of robotics has played a crucial role in research and development. 
The simulation enables robotics application testing to be performed quickly and inexpensively, without being subjected to mechanical or electronic errors, thereby excluding the risk of damage to sensors, motors, and the physical structure of a real robot. 
Virtual reality (VR) is an attractive alternative for interacting with simulated robots because it affords a more immersive experience by  creating better visual cues of real environments. 
Prior research studies illustrate that HRI can be enhanced with VR interfaces \cite{tan2012sigverse} \cite{liu2017understanding}.

\begin{figure}[tbp!]
    \centering
    \includegraphics[width=\linewidth]{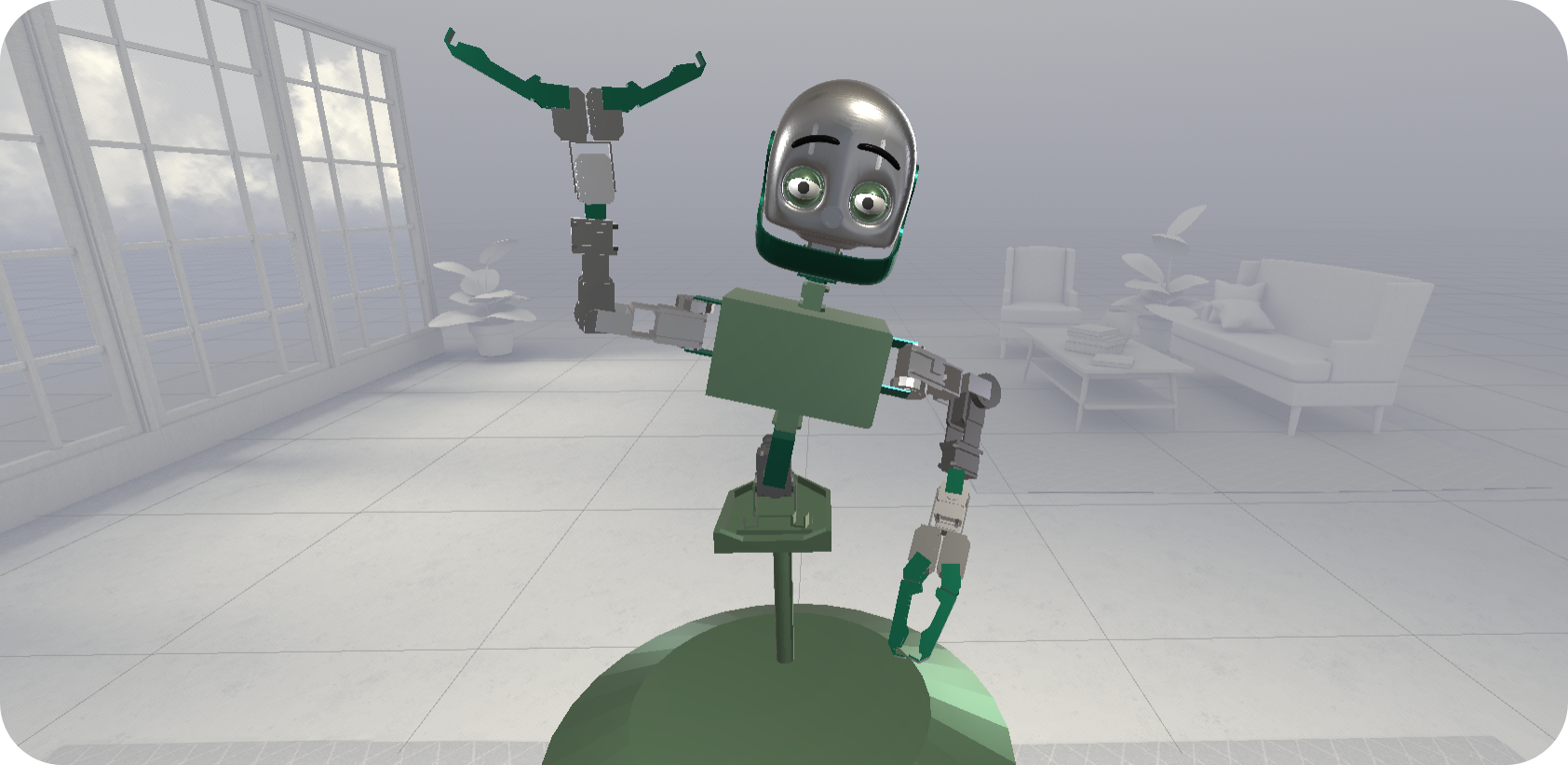}
    \begin{subfigure}[t]{0.11\textwidth}
         \centering
         \includegraphics[width=\textwidth]{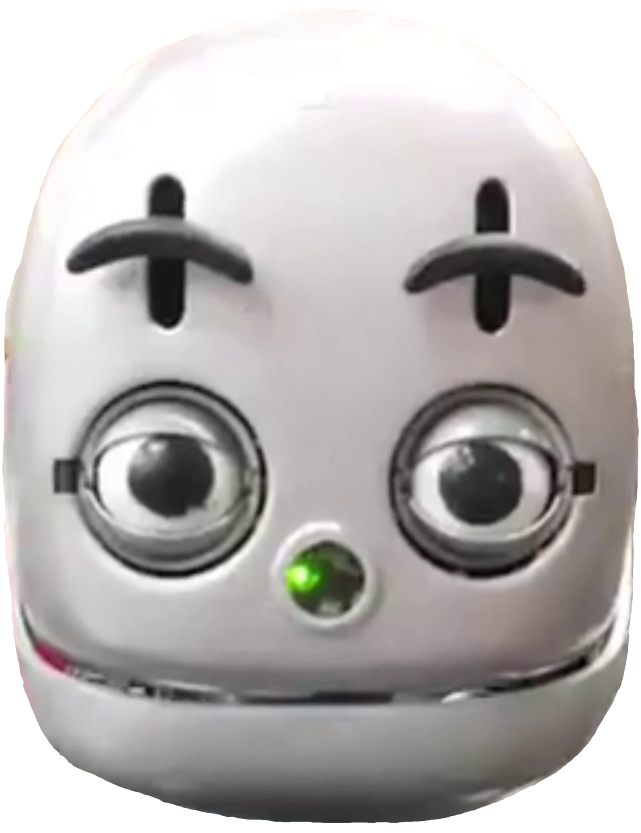}
         \caption{Neutral.}
         \label{fig:idle}
     \end{subfigure}
     \begin{subfigure}[t]{0.11\textwidth}
         \centering
         \includegraphics[width=\textwidth]{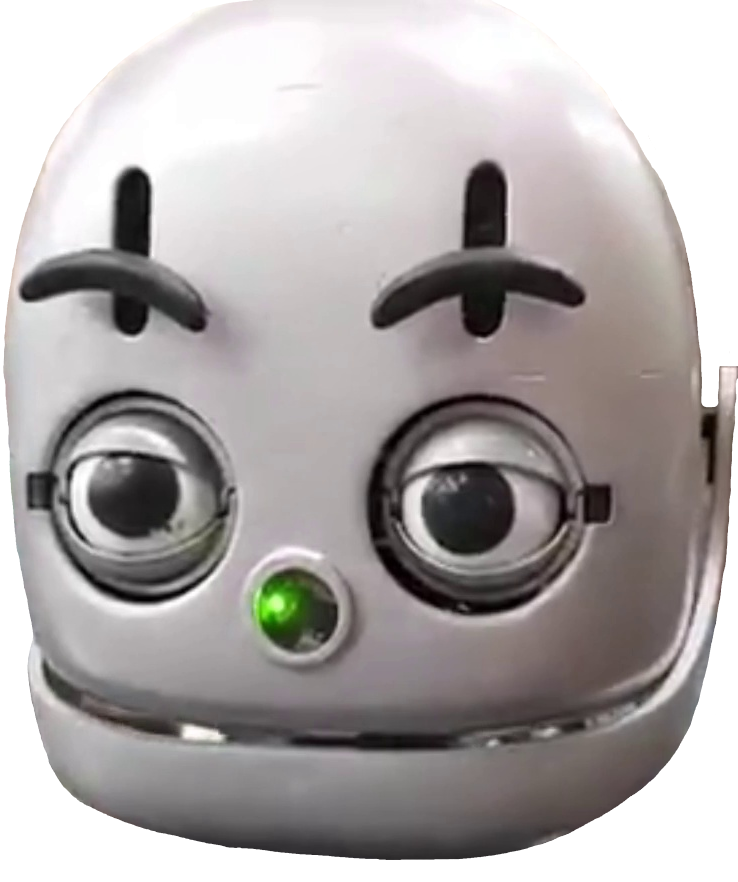}
         \caption{Thinking.} 
         \label{fig:thinking}
     \end{subfigure}
     \begin{subfigure}[t]{0.12\textwidth}
         \centering
         \includegraphics[width=\textwidth]{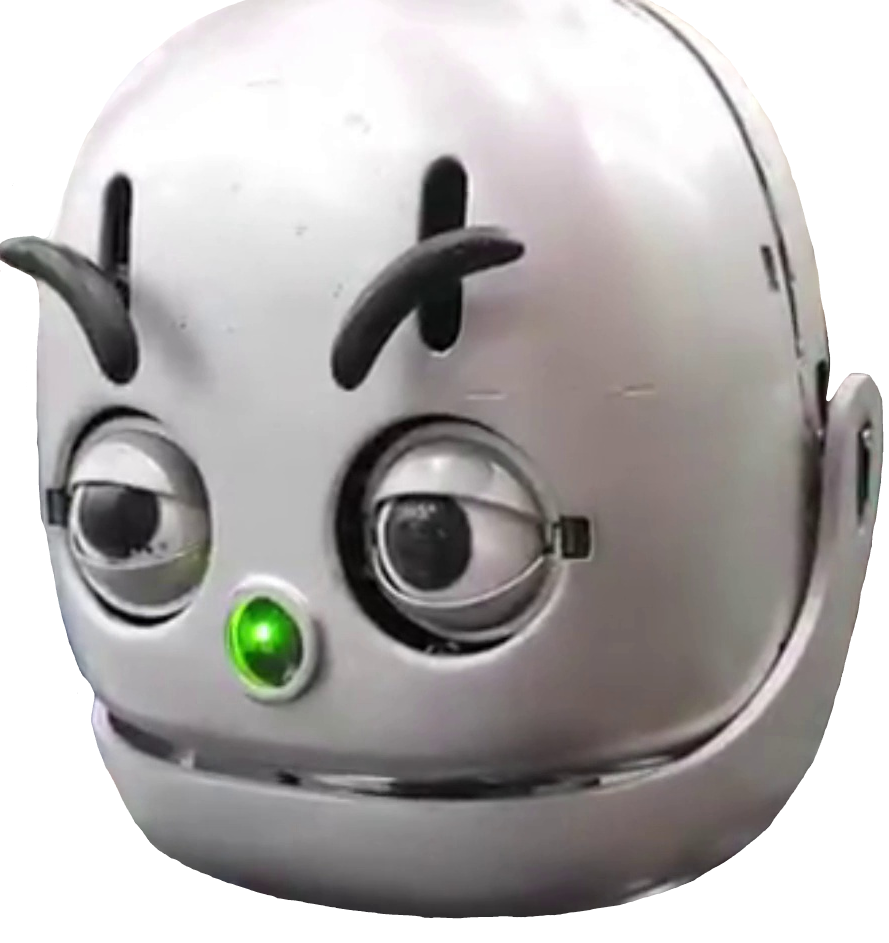}
         \caption{Anger.}
         \label{fig:anger}
     \end{subfigure}
     \begin{subfigure}[t]{0.12\textwidth}
         \centering
         \includegraphics[width=\textwidth]{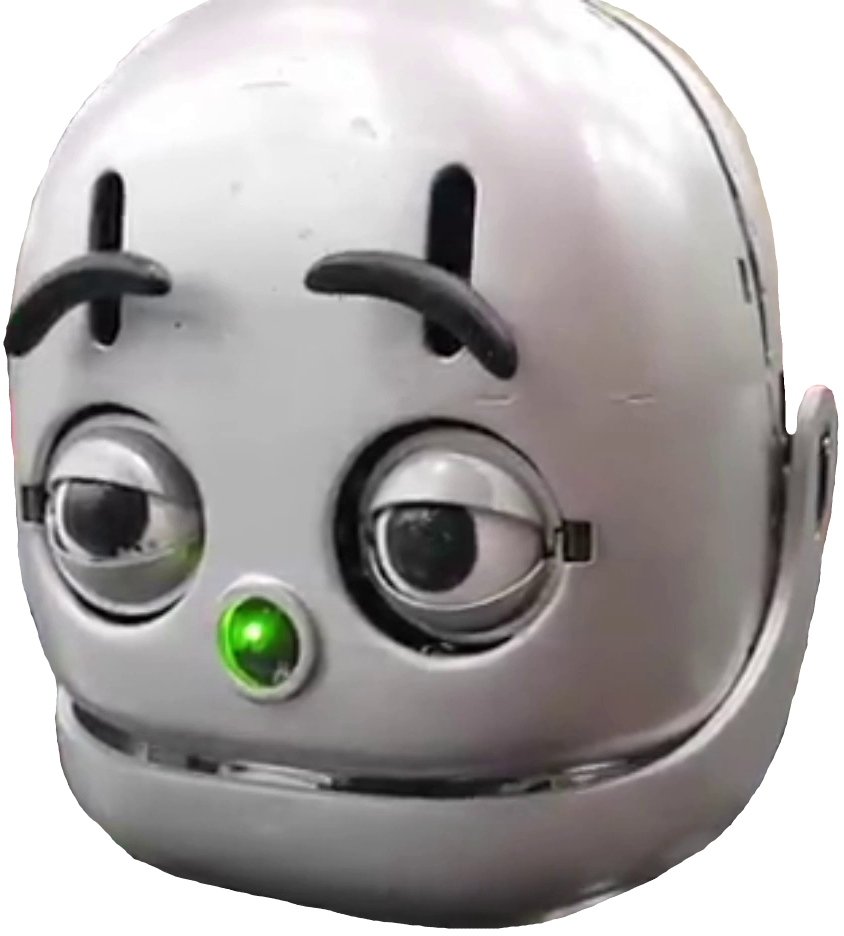}
         \caption{Sadness.}
         \label{fig:sad}
     \end{subfigure}
    \caption{Jubileo's virtual reality framework (up) and Jubileo's animatronic face showing some facial expressions (down).}
    \label{fig:structure}
    \vspace{-5mm}
\end{figure}


The work presented here provides an open-source animatronic face and a VR simulation framework to accomplish immersive experiments. We demonstrate that VR can be an effective way to test simulated social robot applications. Fig.~\ref{fig:structure} shows both the VR framework and Jubileo's physical robot.

This work presents the following contributions:
\begin{itemize}
\item A complete 3D model with its assembly diagram to help the researchers to print their own Jubileo's face using a 3D printer.
\item A low-budget face robot hardware capable of performing many tasks in HRI.
\item An operational system with base functions to HRI social applications.   
\item An open-source framework for development and research in the HRI field using VR.
\end{itemize}

The work is structured as follows: the related works are discussed in Section \ref{related_works}.
In the Section \ref{methodology} 
are explained the systems and techniques used to build the complete research platform of both the physical robot and its framework. The experiments realized are described in Section \ref{experiments}, 
and a brief discussion about the experiments are made in Section \ref{discussion}.
Finally, the main contributions and future works are summarized in Section \ref{conclusions}.        

\section{RELATED WORKS}\label{related_works}


In recent decades, developments in robotics and machine learning have brought robots more and more into human lives and homes. They are no longer limited to simple tasks in the automotive industry, construction, and transportation. With the increase of interactivity between humans and robots emerged a need to study those interactions, originating the robotics research field of HRI. Studying those interactions, Breazeal \cite{breazeal1999context} proposed fundamental properties that a robot should possess to be categorized as a sociable robot.

Considering the expressiveness of the robot, Faraj et al. \cite{face2017faraj}, built a humanoid robotic face, Eva, which can emulate human facial expressions, head movements, and speech through the use of 25 muscles. By creating a face with a great resemblance to a human, Faraj et al created a robot that ended up stepping on the uncanny valley concept.
With the sociability of the robot in mind, Mashiro Mori envisioned the uncanny valley concept \cite{mori2012uncanny}, which depicts the repulsion of humans toward human-like robots. Our robot, Jubileo, has a toy-like face to be more sociable, agreeing with Mori's vision.

Expressions constitute a significant amount of communication between humans \cite{plutchik1984emotions}. So, apart from being aesthetically pleasing, a robotic face ought to be capable of reproducing a variety of expressions and moods. Chen et.al \cite{chen2021smile} proposed an imitation learning method to translate expressions from humans to a humanoid robotic face, based on Faraj's work. These methods, employ deep learning to generate an expression on the robot that imitates human's expressions and controls the face motors. 

The advent of virtual reality provides a means of human-robot interaction without the need for expensive hardware. Chacko et al.\cite{chacko2019augmented}, demonstrated the possibility of interaction with a robot through a cellphone with augmented reality. In this work, we go beyond augmented reality, introducing our robot to an environment set in a virtual reality world, providing total interaction between a human and Jubileo.

Our work differs from the works mentioned above by proposing an uncanny valley-compliant robotic face with a virtual reality framework, enabling human-robot interactions without the need for hardware.

\section{METHODOLOGY}\label{methodology}

In this section, the concept of Jubileo as a research platform is introduced, covering the development details of our approach. We also present our simulated VR scenarios. As well as the necessary hardware to build the physical animatronics face, addressing specific issues such as the communication system.

\begin{figure}[tp!]
    \centering
    \begin{subfigure}[t]{0.23\textwidth}
         \centering
         \includegraphics[width=\textwidth]{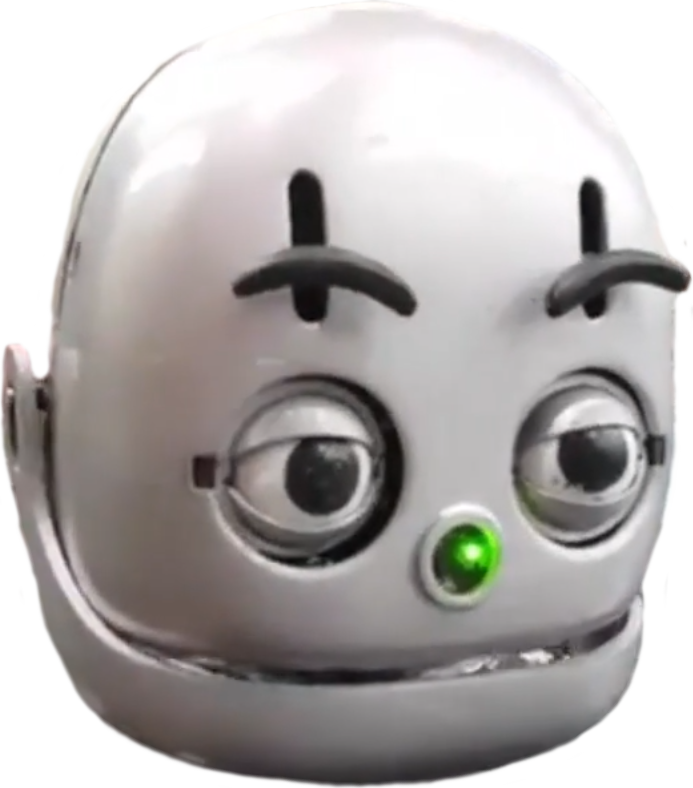}
         \caption{Real.}
         \label{fig:real_face}
     \end{subfigure}
     \begin{subfigure}[t]{0.23\textwidth}
         \centering
         \includegraphics[width=\textwidth]{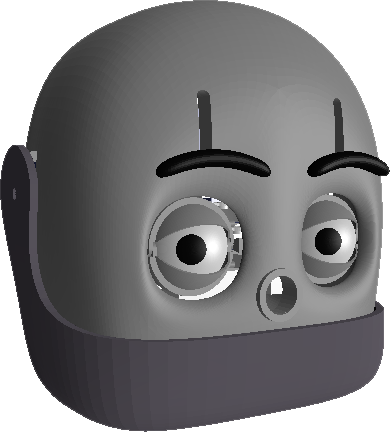}
         \caption{Simulated.}
         \label{fig:simulated_face}
     \end{subfigure}
    \caption{Real and simulated 3D model of Jubileo's face.}
    \label{fig:face}
    \vspace{-5mm}
\end{figure}

\subsection{Animatronic Face}

The facial structure was manufactured in a 3D printer using a material that needed to be strong and resistant, but at the same time, with a good cost-benefit. Considering that, we chose ABS, which fits under these conditions. The robot's face contains $12$ servo-motors, all connected to a microcontroller powered $5$V DC. These servos take care of the movements of the eyes, eyelids, eyebrows, and mouth. The decision-making occurs in a Jetson Nano, sending the necessary data to the microcontroller responsible for receiving and interpreting it.

\begin{table}[bp!]
\vspace{-5mm}
\caption{Specifications of the Jubileo's animatronic face}
\centering
\begin{tabular}{|l|r|}
\hline
\multicolumn{2}{|c|}{Animatronic face components} \\
\hline
Micro Servo SG90 	 	    & 10\\
Micro Servo SG92R           & 2\\
LMS8UU Linear Bearing	    & 2\\
Aluminum Bar (8mm x 70mm)	& 2\\
Arduino Uno			        & 1\\
LM2596 DC 5v		        & 1\\
Arduino Sensor Shield v5.0	& 1\\
Webcam Logitech C310 HD     & 1\\
Rode Compact TRRS Cardioid Mini-Shotgun Microphone & 1\\
Jetson Nano B01             & 1\\
\hline
\multicolumn{2}{|c|}{Physical Specifications} \\
\hline
Height		& 20 cm		\\
Width		& 18 cm		\\
Depth		& 17 cm		\\
Weight		& 0.9 kg	\\
Material	& ABS		\\
\hline
\end{tabular}
\label{table:face_specs}
\end{table}

The Jubileo's design concept proposes a friendly appearance in order to engage people to interact socially, through voice commands, conversations and facial expressions correspondence in an empathetic way. 
This is possible through the use of a speaker for speech synthesis and sensors, being a camera for computer vision and a directional microphone for speech recognition.
Fig.~\ref{fig:face} shows the 3D printed assembled parts and a 3D rendering of the designed face, and Table~\ref{table:face_specs} shows its specifications.

\begin{figure*}[tbp!]
    \centering
     \begin{subfigure}[t]{0.32\textwidth}
         \centering
         \includegraphics[width=\textwidth]{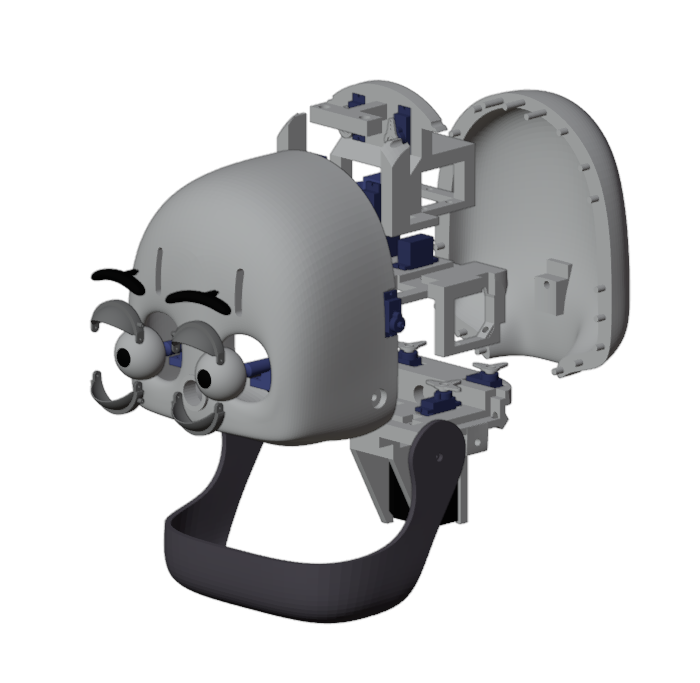}
         \caption{Frontal extruded parts view.}
         \label{fig:extruded_face_front}
     \end{subfigure}
     \begin{subfigure}[t]{0.32\textwidth}
         \centering
         \includegraphics[width=\textwidth]{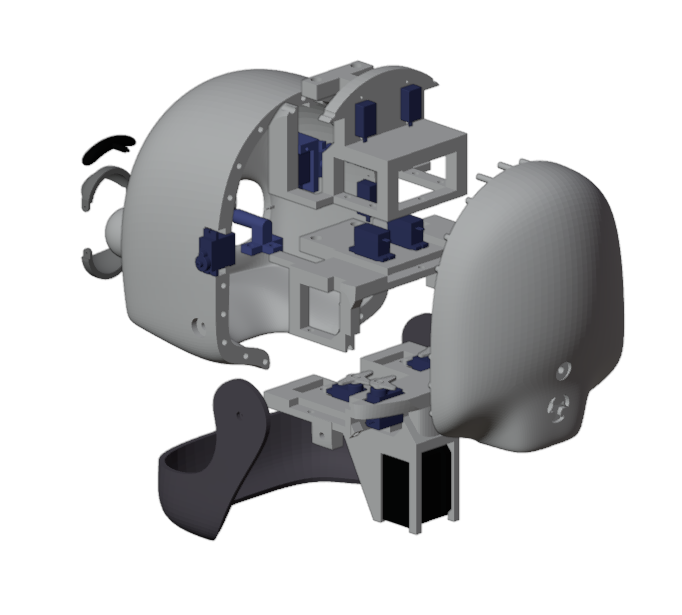}
         \caption{Back extruded parts view.}
         \label{fig:extruded_face_back}
     \end{subfigure}
     \begin{subfigure}[t]{0.34\textwidth}
         \centering
         \includegraphics[width=\textwidth]{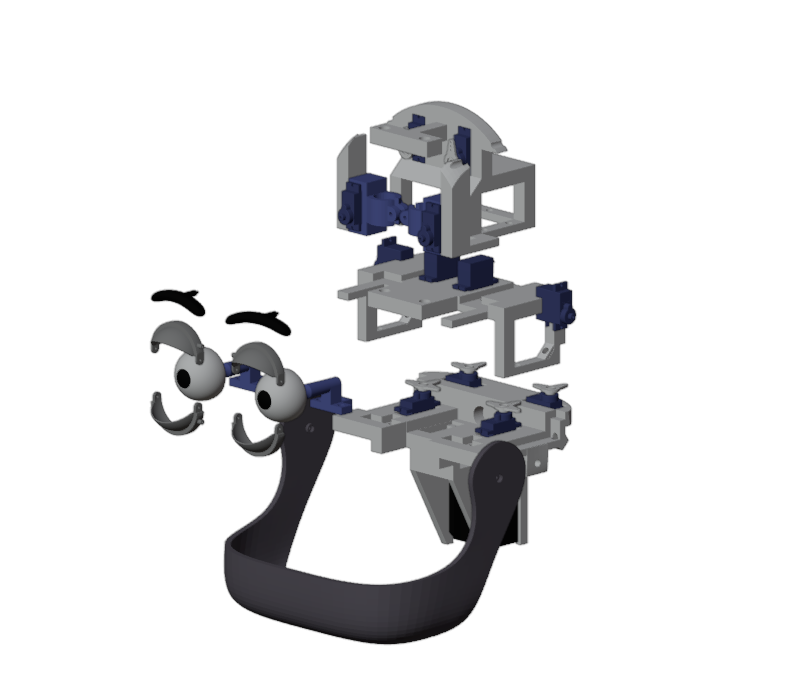}
         \caption{Inner extruded parts view.}
         \label{fig:extruded_face_inner}
     \end{subfigure}
    \caption{Jubileo's 3D model extruded parts. All the mechanisms were built in a modular way to can be switched at any time.}
    \label{fig:extruded_face}
\end{figure*}

The inner 3D structure of our animatronic face model was built to give the maximum interaction possible with its environment through a dynamic demonstration of feelings or attention. Because of that, almost all of the servo motors are used to make the whole movement of eyes, within its eye-brows and eyelids mechanisms, as is shown in Fig.~\ref{fig:extruded_face_front} and Fig.~\ref{fig:extruded_face_back}. The entire 3D model was cropped in a total of 18 parts, so the researcher could print their own Jubileo using any 3D printer.

Two servo motors are used to control the height of the eye-brows at the same time that the others two servo motors are used to control the eye-brows rotation. The horizontal movement of the eyes is produced by the rotational movement of two servo motors attached by small metal rods to the edges of the eyes. In addition to the eyes, vertical movements are performed by two other servo motors using the same concept of horizontal movement. These settings can be seen in Fig.~\ref{fig:extruded_face_inner}.

\subsection{Robot Operating System}

The Robot Operating System (ROS) is a flexible framework for writing software for robots \cite{hart2015affordance}. ROS is a collection of tools and libraries which provides operational system's standard services, like hardware abstraction, low-level device control, messages between processes, and package management. 
Despite the importance of low latency on robot control, ROS is not a real-time operating system, although it is possible to integrate ROS with real-time code. This lack of a real-time system is being addressed in the development of ROS 2 \cite{kay2015real} that is used in this work.

\begin{figure}[bp!]
    \centering
    \includegraphics[width=\linewidth]{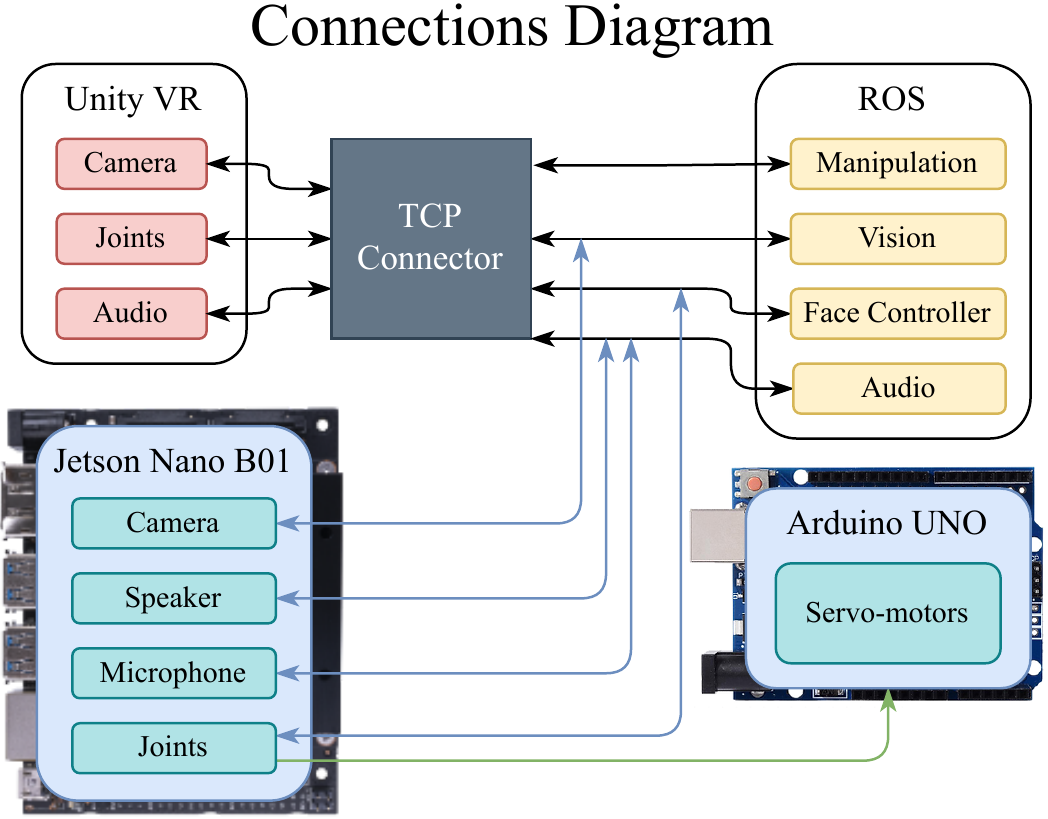}
    \caption{Scheme of ROS connection nodes structure.}
    \label{fig:system_diagram}
\end{figure}

The ROS packages developed for the Jubileo framework aim to provide researchers with a complete base of algorithms that enriches the robot's capabilities, such as computer vision, speech recognition, and robot joint control.
The framework provides a ROS structure that is compatible with both the real and the simulated robot.
A simplified overview of the Jubileo's ROS connection nodes is present in Fig.~\ref{fig:system_diagram}. 

\subsection{Unity and Virtual Reality}

Unity was chosen to make Jubileo's simulated framework, because it is a platform engine that supports the integration with robotics and, at the same time, offers a variety of tools for VR application development \cite{nguyen2017}, in addition to the physics simulation. This platform supports the integration and communication with ROS, and it is possible to import the robot models via the Unified Robotics Description Format (URDF) files \cite{hussein2018}.



The communication between ROS and Unity is made through TCP/IP protocol \cite{meng2015}, a widely used protocol that establishes a connection between client and server before the data could be sent. In this way, the scripts in the Unity platform can exchange information with ROS nodes to control the robot inside the simulation. 
Fig.~\ref{fig:structure} presents the application of the simulated Jubileo attached to a robotic body in Unity, interacting with the user.

Also, we provide the URDF open source models of the robot's face and manipulators, not limiting their use to Unity. 
So, a simulated environment for the Jubileo robot can be done in any desired supported URDF robotics simulation platform.

\begin{figure*}[tbp!]
    \centering
     \begin{subfigure}[t]{0.138\textwidth}
         \centering
         \includegraphics[width=\textwidth]{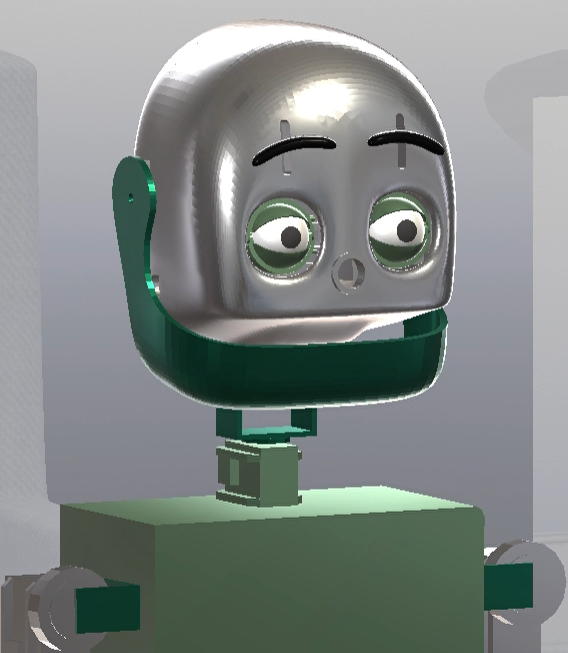}
         \caption{Joy.}
         \label{fig:vr_face_joy}
     \end{subfigure}
     \begin{subfigure}[t]{0.137\textwidth}
         \centering
         \includegraphics[width=\textwidth]{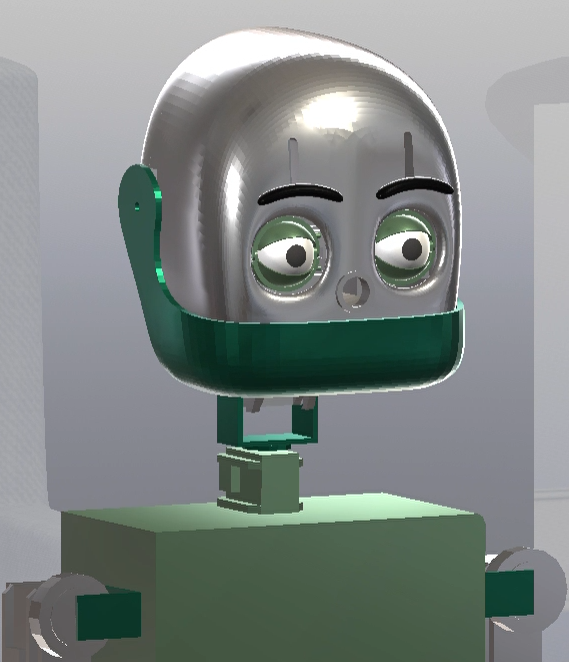}
         \caption{Neutral.}
         \label{fig:vr_face_neutral}
     \end{subfigure}
     \begin{subfigure}[t]{0.137\textwidth}
         \centering
         \includegraphics[width=\textwidth]{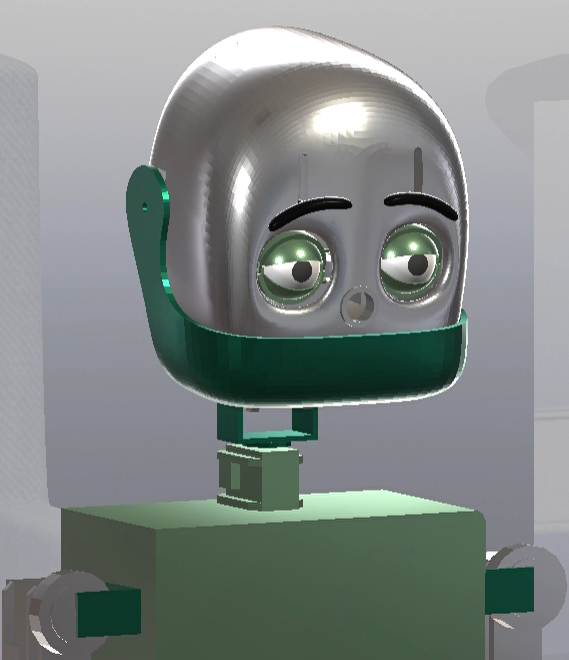}
         \caption{Sadness.}
         \label{fig:vr_face_sadness}
     \end{subfigure}
     \begin{subfigure}[t]{0.138\textwidth}
         \centering
         \includegraphics[width=\textwidth]{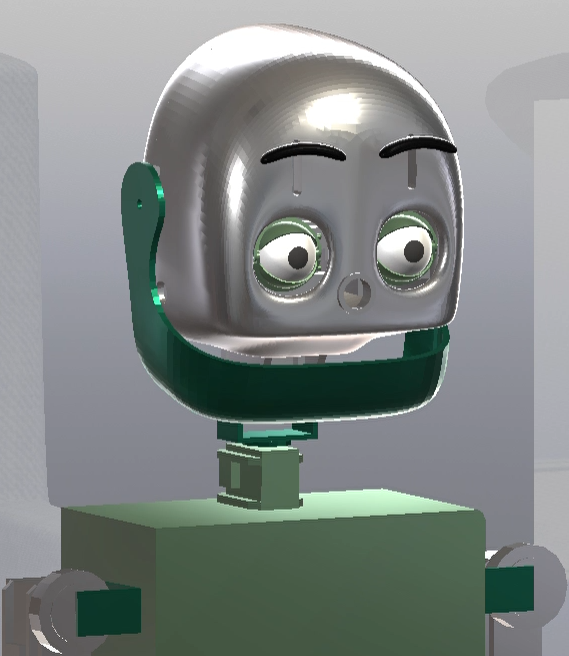}
         \caption{Surprise.}
         \label{fig:vr_face_surprise}
     \end{subfigure}
     \begin{subfigure}[t]{0.136\textwidth}
         \centering
         \includegraphics[width=\textwidth]{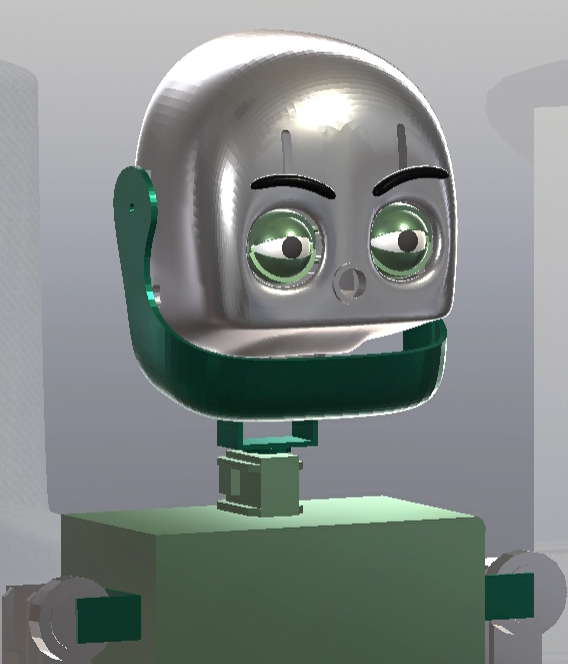}
         \caption{Disgust.}
         \label{fig:vr_face_disgust}
     \end{subfigure}
     \begin{subfigure}[t]{0.138\textwidth}
         \centering
         \includegraphics[width=\textwidth]{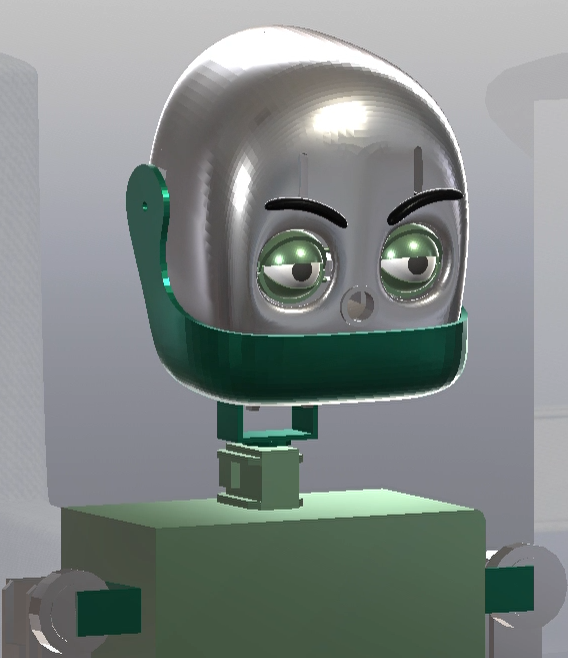}
         \caption{Anger.}
         \label{fig:vr_face_anger}
     \end{subfigure}
     \begin{subfigure}[t]{0.139\textwidth}
         \centering
         \includegraphics[width=\textwidth]{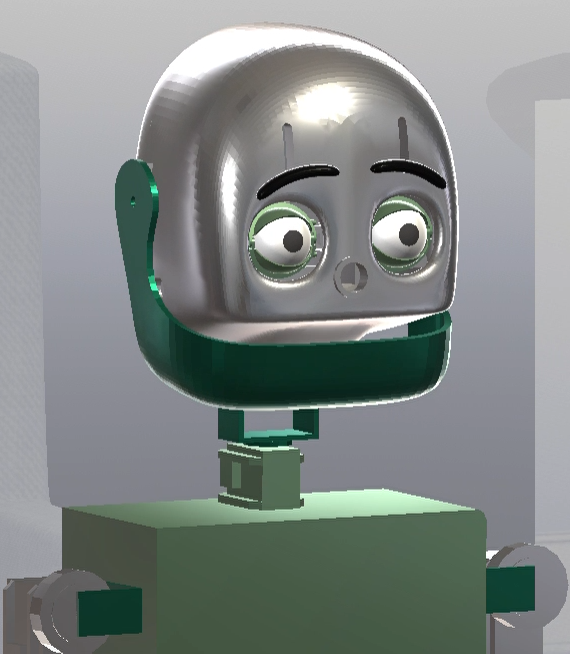}
         \caption{Fear.}
         \label{fig:vr_face_fear}
     \end{subfigure}
    \caption{Facial expressions of Jubileo in simulated environment.}
    \label{fig:vr_face_emotions}
\end{figure*}

\begin{figure*}[bp!]
    \centering
    \includegraphics[width=\textwidth]{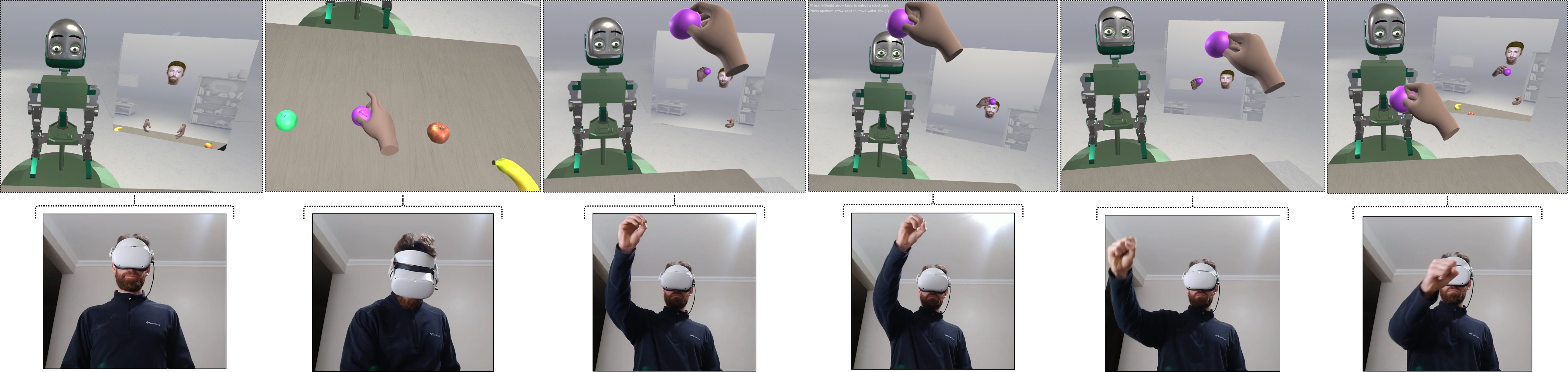}
    \caption{Human object interaction in VR while Jubileo executes visual object tracking.}
    \label{fig:vr_interaction}
\end{figure*}

\section{EXPERIMENTS}\label{experiments}

In this section, we propose some experiments to demonstrate the social capabilities of our robot, interacting with humans and performing facial communication, and other tasks that take a part in the HRI field. Tests were realized in both simulated and real robots.
We kindly invite the interested reader to watch a demonstration of the robot's capabilities at: \url{https://youtu.be/JuxAU4nFGbk}.

\subsection{Facial Expressions and Recognition}\label{facial_expressions}

To better interact and communicate with humans, a robot needs to be able to perform facial expressions and recognize them. In this experiment, we demonstrate the ability of Jubileo to express his mood through his face, as presented in Fig.~\ref{fig:vr_face_emotions}. Our robot also can recognize a human's mood and is capable of imitating it. These skills can be performed in both virtual reality and the real world.

Furthermore, the framework provides a complete face recognition tool.
This utility makes it possible to apply smarter interaction algorithms that can store and use information about specific people.
For example, the robot can remember familiar faces and call the human by name. 

The VR framework platform offers a set of different 3D face models. 
In which, there are a variety of facial expressions and different people. 
The simulated tests in this subsection were performed using some of these 3D avatars.

\begin{figure*}[tbp!]
    \centering
    \includegraphics[width=\textwidth]{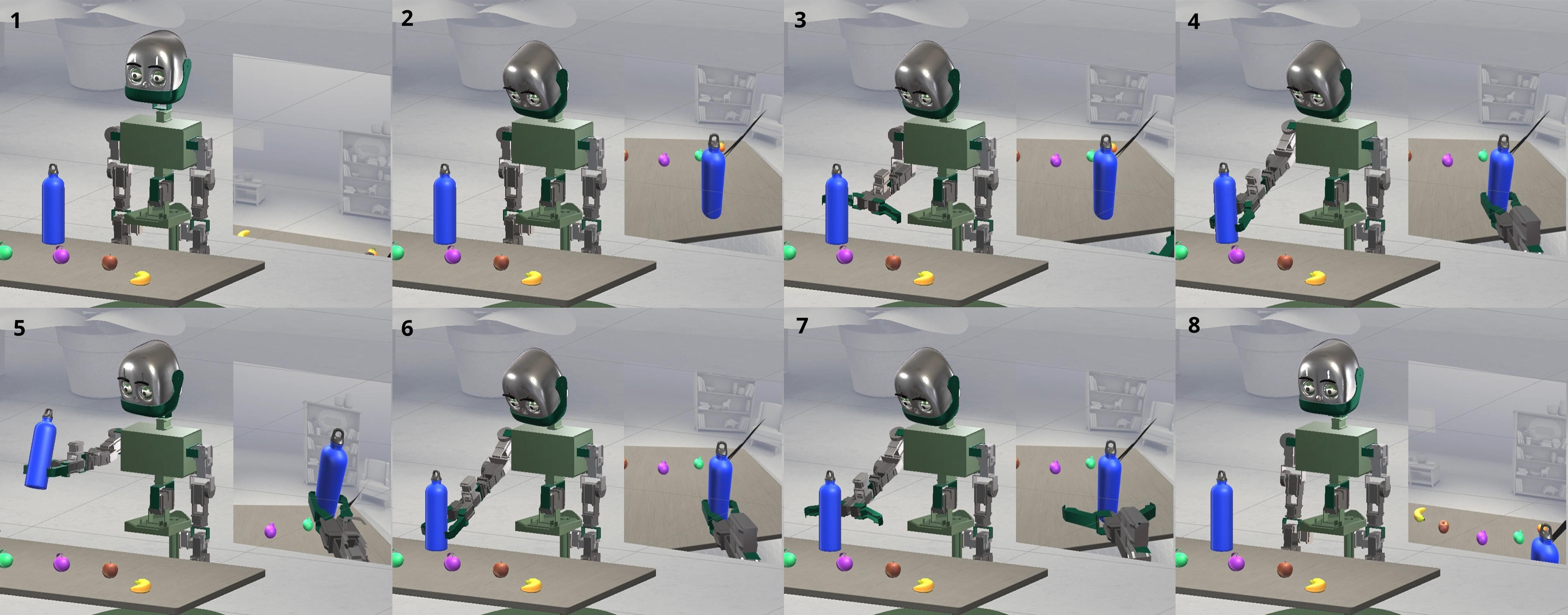}
    \caption{Jubileo grabbing an object in simulated environment.}
    \label{fig:manipulation}
\end{figure*}

\subsection{Visual Object and Face Tracking}\label{visual_tracking}

Visual tracking algorithms are essential for HRI's social applications as they allow the robot to look at people and objects.
Jubileo's framework offers tracking solutions that can be applied through both neck and eyes movement.
The target to be followed is detected by the camera sensor located in the robot's nose hole, and the control of joints are made corresponding to the target position.

In these experiments, the effectiveness of Jubileo's visual tracking in the simulated environment was tested.
At first, we order the robot to look at a particular color.
After he identifies the voice command, a computer vision algorithm detects the object with the specified color.
The human is capable to test the algorithm by interacting with the virtual environment, for example, grabbing and moving the target objects. That interaction can be seen in Fig.~\ref{fig:vr_interaction}. Note that, for better user visualization, a screen was added to the VR environment that demonstrates the robot's camera view.

Secondly, we tested Jubileo's capability of visual face tracking, through face detection.
Giving the voice command "Look at me", the robot starts to gaze at the human.
That is an essential ability for a social robot.

\subsection{Simulated Manipulators}\label{manipulators}

Our simulation framework goes beyond an animatronic face.
Through the simulation, we test the effectiveness of the robotic face attached to a torso with arms, which is capable of making body gestures, as shown in Fig.~\ref{fig:structure}, and more complex tasks — with this, increasing the range of applications that can be developed using the framework.

Therefore, inside the VR environment, in addition to interacting with humans, Jubileo can interact with objects in a virtual reality setting, as demonstrated in Fig.~\ref{fig:manipulation}. This competence was designed so a developer can create and test movement patterns without the real robot, facilitating the development of any application utilizing the robot. In this experiment, it is shown the robot reaching an object employing a series of pre-programmed movements.

\section{DISCUSSION}\label{discussion}

The experiments carried out in the \ref{experiments} session demonstrated the efficiency of our approach in planning and carrying out HRI tasks. However, it is necessary to consider that the developed platform is not restricted to them. As a research development platform, many facilities have been designed to help the researcher build his research more effectively. 
Since the Jubileo framework provides various implemented algorithms, the users do not need to concern with minor code implementations.

Using a simulated environment with a VR as a form of interaction allows HRI researchers to test and validate their algorithms and experiments without the need for complete hardware. In this way, our platform enables the testing of computer vision and reinforcement learning algorithms without requiring a physical system, which significantly speeds up the evaluation process. From this, in addition to the application itself, the researcher can supervise their method and attest their results interactively with the environment through the framework.

The implementation of facial recognition described in the subsection \ref{facial_expressions} and already included in Jubileo guarantees the researcher to act in more targeted interactions, bringing more reliability and immersion to the user interacting with the robot. At the same time, it enables development advances in the order of multi-user interaction. 
The ability to make and recognize facial expressions opens up a range of opportunities in the order of research related to the psychological and behavioral basis.

Furthermore, as shown in the \ref{visual_tracking} subsection, our framework already has implemented visual tracking for objects or human faces. From these tools, it is possible to develop experiments with bias in attention and behavior analysis as well as performing various tests with an emphasis on color-based visual tracking and along with the functionality of easy recognition, attention, and optical monitoring of a person or group of people who are interacting socially with Jubileo. The tracking of the eyes, together with the neck, brings a much more humanized experience to the user, being a researcher or not.

The robotic manipulation system tested in the \ref{manipulators} subsection was implemented in both arms of Jubileo and is easily accessible through the framework. From there, it is possible to carry out experiments by grabbing items and exchanging objects \cite{mathur2022review} between Jubileo and the user. The researcher can quickly improve the manipulation system present in Jubileo by adding image-based pick and place and grasping algorithms \cite{bai2020}. From there, it will only be necessary to communicate the algorithm with the nodes established by the framework. 
In this way, the researcher will have the freedom to explore new manipulation approaches and, consequently, develop more HRI applications.

Finally, the use of voice commands in all previous steps as the primary interaction between the user and the Jubileo robot directs our approach to be used in more research fields. The use of educational robots in teaching sectors has already proven to be efficient in several paradigms \cite{OSPENNIKOVA201518} \cite{anwar2019systematic}. The platform proposed in this work can be used to amplify this type of program by elaborating classes and demonstrations of robotics and programming, with direct interactions between student and robot.
In future work, this can be achieved through the multiplayer approach, where several VR users can enter the same virtual environment, interacting with each other and the robot.

\section{CONCLUSIONS}\label{conclusions}

We present in this work a low-budget open-source robotic face capable of interacting with humans in a manner that agrees with the uncanny valley concept. Together with the animatronic face, we developed a framework in a virtual reality environment enabling interaction with the robot even without the actual hardware. In our experiments, we demonstrate the robot's capabilities to demonstrate his mood, recognize speech and facial expressions, and interact with humans and objects.

Although we demonstrate the ability of Jubileo to interact with humans, this interaction is still constrained to pre-defined situations. Making a more engaging and spontaneous robot is a must to build better social HRI-capable systems. 
Our experiments show that a virtual reality framework is a powerful tool in the construction of HRI systems, making the interaction more meaningful even when interacting with the simulated robot.

Further research is needed to investigate the human perception of Jubileo as a means to produce a more interactive robot.
The ability to understand the context of communication and interaction is still lacking in our research and is a future subject of work.

\section*{Acknowledgement}

The authors would like to thank the VersusAI team. This work was partly founded by the Technological University of Uruguay (UTEC), Federal University of Santa Maria (UFSM), and Federal University of Rio Grande (FURG).

\bibliographystyle{./bibliography/IEEEtran}
\bibliography{./bibliography/IEEEabrv,./bibliography/IEEEreferences}

\begin{thebibliography}{10}
\providecommand{\url}[1]{#1}
\csname url@samestyle\endcsname
\providecommand{\newblock}{\relax}
\providecommand{\bibinfo}[2]{#2}
\providecommand{\BIBentrySTDinterwordspacing}{\spaceskip=0pt\relax}
\providecommand{\BIBentryALTinterwordstretchfactor}{4}
\providecommand{\BIBentryALTinterwordspacing}{\spaceskip=\fontdimen2\font plus
\BIBentryALTinterwordstretchfactor\fontdimen3\font minus
  \fontdimen4\font\relax}
\providecommand{\BIBforeignlanguage}[2]{{%
\expandafter\ifx\csname l@#1\endcsname\relax
\typeout{** WARNING: IEEEtran.bst: No hyphenation pattern has been}%
\typeout{** loaded for the language `#1'. Using the pattern for}%
\typeout{** the default language instead.}%
\else
\language=\csname l@#1\endcsname
\fi
#2}}
\providecommand{\BIBdecl}{\relax}
\BIBdecl

\bibitem{goodrich}
\BIBentryALTinterwordspacing
M.~Goodrich and A.~Schultz, ``Human-robot interaction: A survey,''
  \emph{Foundations and Trends in Human-Computer Interaction}, vol.~1, pp.
  203--275, 01 2007. [Online]. Available:
  \url{https://www.nowpublishers.com/article/Details/HCI-005}
\BIBentrySTDinterwordspacing

\bibitem{plutchik1984emotions}
R.~Plutchik, ``Emotions: A general psychoevolutionary theory,''
  \emph{Approaches to emotion}, vol. 1984, no. 197-219, pp. 2--4, 1984.

\bibitem{mori2012uncanny}
\BIBentryALTinterwordspacing
M.~et~al, ``The uncanny valley [from the field],'' \emph{IEEE Robotics \&
  Automation Magazine}, vol.~19, no.~2, pp. 98--100, 2012. [Online]. Available:
  \url{https://ieeexplore.ieee.org/abstract/document/6213238/}
\BIBentrySTDinterwordspacing

\bibitem{tan2012sigverse}
\BIBentryALTinterwordspacing
J.~T.~C. Tan and T.~Inamura, ``Sigverse-a cloud computing architecture
  simulation platform for social human-robot interaction,'' in \emph{2012 IEEE
  International Conference on Robotics and Automation}.\hskip 1em plus 0.5em
  minus 0.4em\relax IEEE, 2012, pp. 1310--1315. [Online]. Available:
  \url{https://ieeexplore.ieee.org/abstract/document/6225359}
\BIBentrySTDinterwordspacing

\bibitem{liu2017understanding}
\BIBentryALTinterwordspacing
O.~Liu, D.~Rakita, B.~Mutlu, and M.~Gleicher, ``Understanding human-robot
  interaction in virtual reality,'' in \emph{2017 26th IEEE international
  symposium on robot and human interactive communication (RO-MAN)}.\hskip 1em
  plus 0.5em minus 0.4em\relax IEEE, 2017, pp. 751--757. [Online]. Available:
  \url{https://ieeexplore.ieee.org/abstract/document/8172387}
\BIBentrySTDinterwordspacing

\bibitem{breazeal1999context}
\BIBentryALTinterwordspacing
C.~Breazeal and B.~Scassellati, ``A context-dependent attention system for a
  social robot,'' \emph{rn}, vol. 255, p.~3, 1999. [Online]. Available:
  \url{https://www.cs.ryerson.ca/aferworn/courses/CPS841/CLASSES/CPS841CL05/Breazeal-Scaz-IJCAI99.pdf}
\BIBentrySTDinterwordspacing

\bibitem{face2017faraj}
\BIBentryALTinterwordspacing
Z.~Faraj, M.~Selamet, C.~Morales, P.~Torres, M.~Hossain, B.~Chen, and
  H.~Lipson, ``Facially expressive humanoid robotic face,'' \emph{HardwareX},
  vol.~9, p. e00117, 2021. [Online]. Available:
  \url{https://www.sciencedirect.com/science/article/pii/S2468067220300262}
\BIBentrySTDinterwordspacing

\bibitem{chen2021smile}
\BIBentryALTinterwordspacing
B.~Chen, Y.~Hu, L.~Li, S.~Cummings, and H.~Lipson, ``Smile like you mean it:
  Driving animatronic robotic face with learned models,'' in \emph{2021 IEEE
  International Conference on Robotics and Automation (ICRA)}.\hskip 1em plus
  0.5em minus 0.4em\relax IEEE, 2021, pp. 2739--2746. [Online]. Available:
  \url{https://ieeexplore.ieee.org/abstract/document/9560797}
\BIBentrySTDinterwordspacing

\bibitem{chacko2019augmented}
\BIBentryALTinterwordspacing
S.~M. Chacko and V.~Kapila, ``An augmented reality interface for human-robot
  interaction in unconstrained environments,'' in \emph{2019 IEEE/RSJ
  International Conference on Intelligent Robots and Systems (IROS)}.\hskip 1em
  plus 0.5em minus 0.4em\relax IEEE, 2019, pp. 3222--3228. [Online]. Available:
  \url{https://ieeexplore.ieee.org/abstract/document/8967973}
\BIBentrySTDinterwordspacing

\bibitem{hart2015affordance}
\BIBentryALTinterwordspacing
S.~Hart, P.~Dinh, and K.~Hambuchen, ``The affordance template ros package for
  robot task programming,'' in \emph{2015 IEEE international conference on
  robotics and automation (ICRA)}.\hskip 1em plus 0.5em minus 0.4em\relax IEEE,
  2015, pp. 6227--6234. [Online]. Available:
  \url{https://ieeexplore.ieee.org/abstract/document/7140073}
\BIBentrySTDinterwordspacing

\bibitem{kay2015real}
\BIBentryALTinterwordspacing
J.~Kay and A.~R. Tsouroukdissian, ``Real-time control in ros and ros 2.0,''
  \emph{ROSCon15}, 2015. [Online]. Available:
  \url{https://www.osrfoundation.org/wordpress2/wp-content/uploads/2015/11/ROSCon15-Kay.pdf}
\BIBentrySTDinterwordspacing

\bibitem{nguyen2017}
\BIBentryALTinterwordspacing
V.~T. Nguyen and T.~Dang, ``Setting up virtual reality and augmented reality
  learning environment in unity,'' in \emph{2017 IEEE International Symposium
  on Mixed and Augmented Reality (ISMAR-Adjunct)}, 2017, pp. 315--320.
  [Online]. Available:
  \url{https://ieeexplore.ieee.org/abstract/document/8088512}
\BIBentrySTDinterwordspacing

\bibitem{hussein2018}
\BIBentryALTinterwordspacing
A.~Hussein, F.~García, and C.~Olaverri-Monreal, ``Ros and unity based
  framework for intelligent vehicles control and simulation,'' in \emph{2018
  IEEE International Conference on Vehicular Electronics and Safety (ICVES)},
  2018, pp. 1--6. [Online]. Available:
  \url{https://ieeexplore.ieee.org/abstract/document/8519522}
\BIBentrySTDinterwordspacing

\bibitem{meng2015}
\BIBentryALTinterwordspacing
W.~Meng, Y.~Hu, J.~Lin, F.~Lin, and R.~Teo, ``Ros+unity: An efficient
  high-fidelity 3d multi-uav navigation and control simulator in gps-denied
  environments,'' in \emph{IECON 2015 - 41st Annual Conference of the IEEE
  Industrial Electronics Society}, 2015, pp. 002\,562--002\,567. [Online].
  Available: \url{https://ieeexplore.ieee.org/abstract/document/7392488}
\BIBentrySTDinterwordspacing

\bibitem{mathur2022review}
\BIBentryALTinterwordspacing
A.~Mathur, C.~Bansal, S.~Chauhan, and O.~Yadav, ``A review of pick and place
  operation using computer vision and ros,'' \emph{Computational and
  Experimental Methods in Mechanical Engineering}, pp. 411--418, 2022.
  [Online]. Available:
  \url{https://link.springer.com/chapter/10.1007/978-981-16-2857-3_41}
\BIBentrySTDinterwordspacing

\bibitem{bai2020}
\BIBentryALTinterwordspacing
Q.~Bai, S.~Li, J.~Yang, Q.~Song, Z.~Li, and X.~Zhang, ``Object detection
  recognition and robot grasping based on machine learning: A survey,''
  \emph{IEEE Access}, vol.~8, pp. 181\,855--181\,879, 2020. [Online].
  Available: \url{https://ieeexplore.ieee.org/abstract/document/9212350}
\BIBentrySTDinterwordspacing

\bibitem{OSPENNIKOVA201518}
\BIBentryALTinterwordspacing
E.~Ospennikova, M.~Ershov, and I.~Iljin, ``Educational robotics as an inovative
  educational technology,'' \emph{Procedia - Social and Behavioral Sciences},
  vol. 214, pp. 18--26, 2015, worldwide trends in the development of education
  and academic research, Sofia, Bulgaria, 15-18 June, 2015. [Online].
  Available:
  \url{https://www.sciencedirect.com/science/article/pii/S1877042815059431}
\BIBentrySTDinterwordspacing

\bibitem{anwar2019systematic}
\BIBentryALTinterwordspacing
S.~Anwar, N.~A. Bascou, M.~Menekse, and A.~Kardgar, ``A systematic review of
  studies on educational robotics,'' \emph{Journal of Pre-College Engineering
  Education Research (J-PEER)}, vol.~9, no.~2, p.~2, 2019. [Online]. Available:
  \url{https://docs.lib.purdue.edu/jpeer/vol9/iss2/2/}
\BIBentrySTDinterwordspacing

\end{thebibliography}

\addtolength{\textheight}{-12cm}  

\end{document}